\documentclass[preprint,12pt]{elsarticle}

\usepackage{amssymb}
\usepackage{amsmath}
\usepackage{listings}
\usepackage{booktabs}
\usepackage{wrapfig}

\usepackage{url}

\usepackage{tcolorbox}
\definecolor{indigo}{rgb}{0.29, 0.0, 0.51}
\definecolor{calpolypomonagreen}{rgb}{0.12, 0.3, 0.17}
\definecolor{uclagold}{rgb}{1.0, 0.7, 0.0}

\newcommand{\proposed}{{AutoPrognosis}}
\newcommand{\proposedf}{{AutoPrognosis 2.0}}
\begin{document}

\begin{frontmatter}



\title{{\proposedf}: Democratizing Diagnostic and Prognostic Modeling in Healthcare with Automated Machine Learning}


\author[inst1]{Fergus Imrie}
\author[inst2]{Bogdan Cebere}
\author[inst3]{Eoin F. McKinney}
\author[inst2,inst4]{Mihaela van der Schaar}

\affiliation[inst1]{organization={Department of Electrical and Computer Engineering, University of California},%Department and Organization
            city={Los Angeles},
            state={CA},
            country={USA}}

\affiliation[inst2]{organization={Department of Applied Mathematics and Theoretical Physics, University of Cambridge},%Department and Organization
            city={Cambridge},
            country={UK}}

\affiliation[inst3]{organization={Department of Medicine, University of Cambridge},%Department and Organization
            city={Cambridge},
            country={UK}}

\affiliation[inst4]{organization={The Alan Turing Institute},%Department and Organization
            city={London},
            country={UK}}

%\affiliation[inst6]{organization={Department of Public Health and Primary Care, University of Cambridge},%Department and Organization
%            city={Cambridge},
%            country={UK}}

\begin{abstract}
Diagnostic and prognostic models are increasingly important in medicine and inform many clinical decisions.
Recently, machine learning approaches have shown improvement over conventional modeling techniques by better capturing complex interactions between patient covariates in a data-driven manner. 
However, the use of machine learning introduces a number of technical and practical challenges that have thus far restricted widespread adoption of such techniques in clinical settings. 
To address these challenges and empower healthcare professionals, we present a machine learning framework, {\proposedf}, to develop diagnostic and prognostic models.
{\proposed} leverages state-of-the-art advances in automated machine learning to develop optimized machine learning pipelines, incorporates model explainability tools, and enables deployment of clinical demonstrators, \textit{without} requiring significant technical expertise.
Our framework eliminates the major technical obstacles to predictive modeling with machine learning that currently impede clinical adoption. 
To demonstrate {\proposedf}, we provide an illustrative application where we construct a prognostic risk score for diabetes using the UK Biobank, a prospective study of 502,467 individuals.
The models produced by our automated framework achieve greater discrimination for diabetes than expert clinical risk scores.
Our risk score has been implemented as a web-based decision support tool\footnote{\url{https://autoprognosis-biobank-diabetes.streamlitapp.com/}} and can be publicly accessed by patients and clinicians worldwide.
In addition, {\proposedf} is provided as an open-source python package.
By open-sourcing our framework as a tool for the community, clinicians and other medical practitioners will be able to readily develop new risk scores, personalized diagnostics, and prognostics using modern machine learning techniques.

\vspace{\baselineskip}
\noindent \textbf{Software:} \url{https://github.com/vanderschaarlab/AutoPrognosis}
\end{abstract}

\end{frontmatter}


\section{Introduction}
\label{sec:intro}

Machine learning (ML) systems have the potential to revolutionize medicine and become core clinical tools \cite{Topol2019}. 
However, there are a diverse set of challenges that must be overcome prior to routine and widespread ML adoption \cite{Gerke2020,Sun2019}. 
In particular, there are substantial technical challenges in developing, understanding, and deploying ML systems which currently render them largely inaccessible for medical practitioners \cite{Sun2019,Yu2018,Rajpurkar2022,Petersson2022}.

In an attempt to address this, we previously developed {\proposed}, an automated machine learning (AutoML) framework to train predictive models \cite{Alaa2018}.
This framework has since been applied to derive prognostic models for cardiovascular disease \cite{Alaa2019}, cystic fibrosis \cite{Alaa2018CF}, and breast cancer \cite{Alaa2021}, among a number of other indications \cite{Rahbar2020,Qian2021,Shah2021THA,Devana2021TKA,Shah2022spinalfusion,Shah2022C5}.
However, our initial approach had significant limitations from both algorithmic and usability perspectives.  

Consequently, in this work, we describe {\proposedf}, which addresses all major obstacles limiting the development, interpretation and deployment of ML methods in medicine and represents a step-change in diagnostic and prognostic modeling. 
In particular, we believe this is the world's first method that can simultaneously: 
(1) solve classification, regression, and time-to-event problems; 
(2) optimize ML pipelines, determine the most appropriate models, and automatically tune hyperparameters;
(3) identify key variables and novel risk factors, enabling clinicians to select different numbers of variables and understand the value of information;
(4) provide a diverse range of model explanations, including feature-based, example-based, and closed-form risk equations;
and (5) produce web-based applications, allowing models to be readily shared with the clinical community.

In this paper, we outline the major challenges facing clinical development and translation of diagnostic and prognostic modeling. 
We then describe our approach, {\proposedf}, and detail how it addresses each challenge. 
Finally, we demonstrate the application of {\proposedf} in an illustrative scenario: prognostic risk prediction of diabetes using a cohort of 502,467 individuals from UK Biobank.
However, we emphasize that {\proposed} can be applied to construct diagnostic and prognostic models for \textit{any} disease or clinical outcome, and is explicitly designed to make model building accessibly by non-ML experts.
We have open-sourced {\proposedf} as a tool for the community, allowing clinicians or non-expert users to adopt the automated framework to robustly and reproducibly develop optimized personalized diagnostics, prognostics, and risk scores using modern machine learning techniques.

\section{Challenges in Diagnostic and Prognostic Modeling}
\label{sec:challenges}

There are numerous obstacles to developing and deploying diagnostic and prognostic models that currently prevent healthcare professionals from capitalizing on recent algorithmic advances \cite{Topol2019}.
Our work seeks to empower clinicians, medical researchers, epidemiologists, and biostatisticians through an accessible, automated framework capable of identifying optimal solutions to all major obstacles limiting ML model building with minimal need for technical expertise.
We begin by describing the seven major challenges faced by these communities and how they are addressed by {\proposedf}.

\begin{table}
    \centering
    \begin{tabular}{c}
    \begin{tcolorbox}[width=\textwidth,colback={red!5},title={Challenge 1. Developing powerful ML pipelines},colframe=red!75!black] 
    {\proposed} uses AutoML to automate pipeline configuration, performing missing value imputation, feature processing, model selection, and hyperparameter optimization.
    \end{tcolorbox} 
    \\
    \begin{tcolorbox}[width=\textwidth,colback={orange!5},title={Challenge 2. Understanding the value of ML and when it is necessary},colframe=orange!75!black] 
    {\proposed} compares a range of ML methods to traditional approaches and automatically identifies what approach is best.
    \end{tcolorbox} 
    \\
    \begin{tcolorbox}[width=\textwidth,colback={uclagold!5},title={Challenge 3. Determining the value of information},colframe=uclagold!75!black] 
    {\proposed} can quantify the value of including additional predictors, enabling systematic identification of optimal variables.
    \end{tcolorbox} 
    \\
    \begin{tcolorbox}[width=\textwidth,colback={calpolypomonagreen!5},title={Challenge 4. Understanding and debugging ML models},colframe=calpolypomonagreen!75!black] 
    {\proposed} incorporates seven state-of-the-art interpretability methods, allowing models to be understood and debugged as they are generated.
    \end{tcolorbox} 
    \\
    \begin{tcolorbox}[width=\textwidth,colback={blue!5},title={Challenge 5. Making ML models accessible and usable},colframe=blue!75!black] 
    {\proposed} provides a platform to share model outputs by automating the creation of web-based applications. 
    \end{tcolorbox} 
    \\
    \begin{tcolorbox}[width=\textwidth,colback={indigo!5},title={Challenge 6. Deciding when and if to update clinical models},colframe=indigo!75!black] 
    {\proposed} can quantify the benefit of additional data or new predictive variables, and automatically determine the optimal system for the new dataset. 
    \end{tcolorbox} 
    \\
    \begin{tcolorbox}[width=\textwidth,colback={violet!5},title={Challenge 7. Transparent reproducibility},colframe=violet!75!black] 
    {\proposed} provides a standardized, publicly available framework, facilitating reproducibility.
    \end{tcolorbox} 
    \\
    \end{tabular}
    \caption{Major challenges facing clinical development of diagnostic and prognostic models and how these are addressed by {\proposed}. See Section \ref{sec:challenges} for more detail.}
    \label{tab:my_label}
\end{table}

\subsection*{Challenge 1. Developing powerful ML pipelines}

%probst2019tunability
Developing performant ML models remains complex and typically involves significant time and effort, even for expert ML practitioners. Indeed, some estimates suggest over 95\% of work is expended on software technicals, leaving less than 5\% for addressing the medical or scientific problem at hand \cite{sculley2015hidden}. 
This is further complicated by the myriad of choices that must be made when developing a new predictive model for diagnosis or prognosis, 
such as: what imputation strategy should be used; how should the data be preprocessed; what (ML) model is best suited for the specific task; what configuration of hyperparameters should be used. 
These decisions affect each other, thus cannot be made in isolation; further, the optimal choices not only vary between applications, but also can change over time as more data is collected and clinical practice changes \cite{nestor2018rethinking}.

Few resources are available to help empirically define optimal computational pipelines.
{\proposedf} addresses this by incorporating an AutoML approach within a standardized framework, automating the process of pipeline configuration. 
{\proposed} navigates a broad algorithmic search space in an efficient fashion, systematically performing missing value imputation, feature processing, model selection, and hyperparameter optimization in an unbiased manner without the need for human intervention or expert insight. This avoids arbitrary parameter selection and ensures standardization of pipelines, facilitating both reproducibility and optimized model performance. 
Critically, this democratizes the model building step, eliminating the requirement for expert ML knowledge and making cutting-edge methodology accessible to all, freeing healthcare domain experts to define and address the core clinical problems.

\subsection*{Challenge 2. Understanding the value of ML and when it is necessary}

Traditional approaches, such as linear regression and Cox proportional hazard models \cite{Cox1972}, are widely used and accepted across healthcare.
Before replacing these established methods, it is vital to understand whether ML is valuable for a given problem and quantify the benefit of ML systems. Indeed, there is no ``free lunch'' and we should not expect ML to always outperform existing approaches.
Several recent examples exist that present settings where comparatively ``simple'' approaches outperformed ML \cite{AKBILGIC2019, Schulz2020}.
{\proposedf} can be used to compare a range of ML methods to traditional approaches at minimal technical cost to the user.  
Furthermore, since these solutions are included in the algorithmic search space, {\proposed} will automatically identify whether such approaches are indeed best or if more complex ML models are required. 

\subsection*{Challenge 3. Determining the value of information}

Selecting which variables to include in a predictive model represents a key decision that not only impacts model performance but also the ease of subsequent clinical use since any feature used will need to be collected in an ongoing manner to use such systems.
Thus, understanding the \textit{value} of an individual variable and the information it provides is critical. Often, this is assessed by univariate statistical analysis or other selection methods such as forward selection or backwards elimination \cite{guyon2003introduction}.
{\proposedf} provides methods to test and quantify the value of including additional predictors, allowing systematic identification of optimal variables in an informed manner.

\subsection*{Challenge 4. Understanding and debugging ML models}

A predictive clinical model must be more than just accurate, it must be interpretable. Without a transparent understanding of \textit{how} a model makes predictions it may act in unintended and undesirable ways, for example learning incorrect or aberrant features unique to the training data \cite{caruana2015intelligible,Winkler2019}. 
This debugging step is critical for building model trust \cite{Rajpurkar2022} and cannot be achieved without interpretation of the training features or cases that support model accuracy.
It is clear that clinical deployment of an interpretable model is supported by the additional trust gained by understanding the models performance \cite{Yoon2022}.

Furthermore, a clear understanding of computational models is now a requirement for deployment in healthcare systems globally: in the United States, the FDA demands ``transparency about the function and modifications of medical devices'' as a key safety aspect \cite{FDA2019}, while Article 22 of GDPR legislation in the EU requires that ``meaningful information about the logic involved'' be provided \cite{Mourby2021}.
To achieve this transparency, interpretable outputs of a specific form are typically required. For example, the American Joint Committee on Cancer requires explicit risk equations \cite{Kattan2016}.
The `black-box' nature of many ML methods means that they remain inherently uninterpretable and require specialized methods to unravel the underlying rationale for predictions.
In {\proposedf}, we have incorporated seven state-of-the-art interpretability methods allowing researchers to understand and debug ML models as they are generated. 

\subsection*{Challenge 5. Making ML models accessible and usable}

Predictive models need to be accessible to be used in clinical practice.
This step often limits adoption, since bespoke deployment can result in significant costs and reliance on technical expertise. 
While full clinical deployment may require additional systems (e.g. due to regulatory requirements), a standardized, user-friendly solution to rapidly visualize and share models is also a necessary part of both debugging and confirming clinical acceptance.
{\proposedf} provides a platform to share model outputs by automating the creation of web-based applications, allowing clinicians to explore predictions in diverse scenarios. 

\subsection*{Challenge 6. Deciding when and if to update clinical models}

Over time, more data is collected, new variables are measured, and even clinical practice changes. For the former, existing clinical predictive models might benefit from additional data or features, while in the latter case, model performance may degrade \cite{nestor2018rethinking}. 
However, deciding whether to update a clinical model is not a decision to be made lightly, since beyond model building, further regulatory approval might be necessary and the updated model will need to be redeployed.
{\proposed} can help answer this difficult question by quantifying the benefit of additional data and new predictive variables, while also automatically determining the optimal system configurations for the new dataset, which may have changed. 

\subsection*{Challenge 7. Transparent reproducibility}

Reproducibility is a fundamental requirement for the acceptance and adoption of any predictive model. 
While transparently reproducing a model's output on a given dataset is conceptually simple, several factors can confound this necessary step. Serial data releases, code updates and even inherent properties of ML algorithms (for example, stochastic descent methods can give different answers even when run repeatedly on the same data) can conspire to make ML model building less reproducible than it should be \cite{Beam2020}.
These issues demonstrably obstruct translation of clinical prediction and erode trust in ML approaches \cite{LeVeque2012,Milkowski2018}. 
{\proposedf} addresses this major challenge by providing a standardized, publicly available framework to train predictive models, allowing straightforward demonstration of reproducibility on source data.

\begin{figure*}[ht]%
\centering
\includegraphics[trim={0 0.75cm 0 0.75cm}, clip, width=\textwidth]{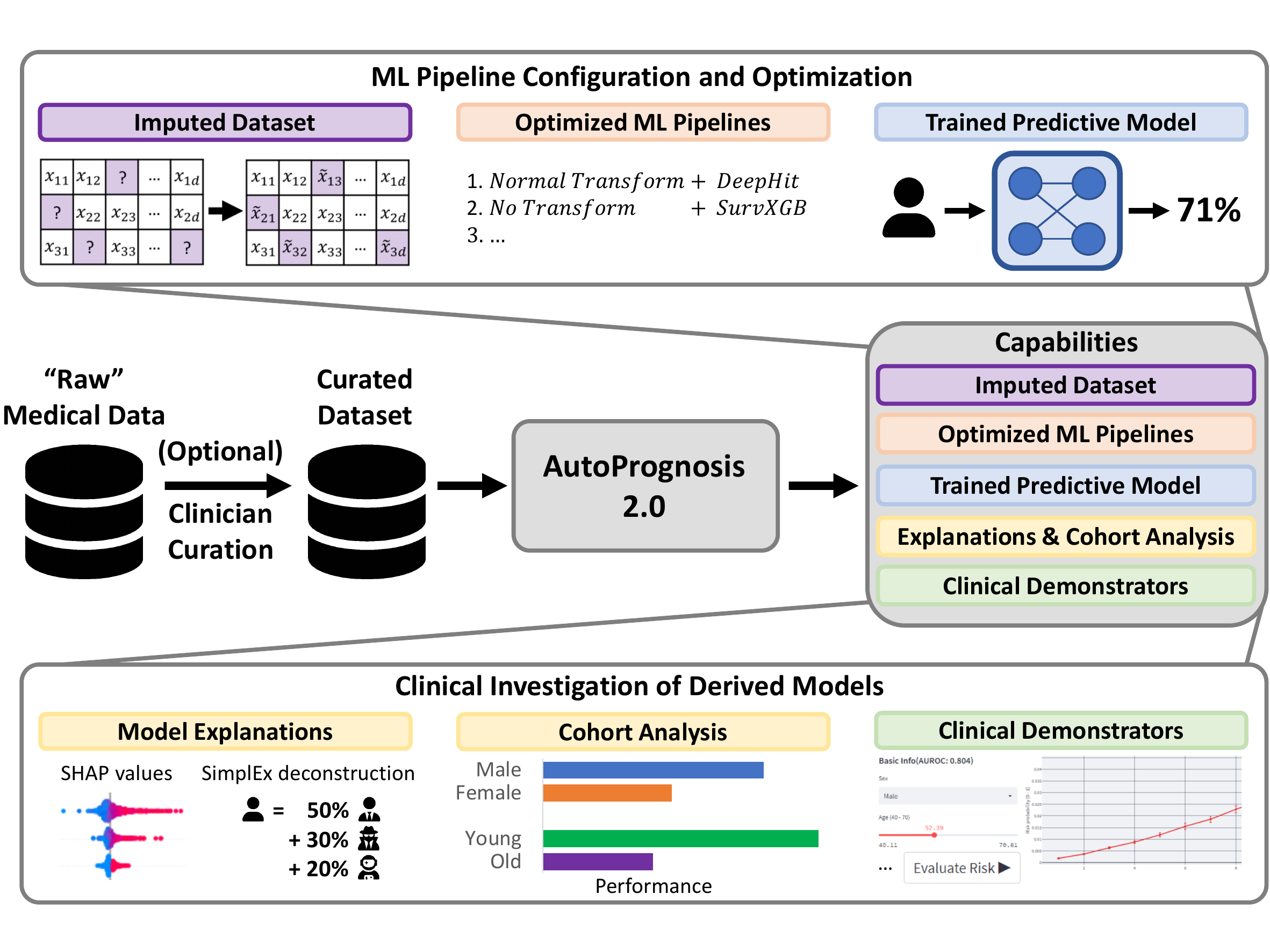}
\caption{Overview of the {\proposedf} framework. {\proposed} takes either raw or curated medical datasets and provides an imputed dataset, a report detailing the optimized machine learning pipelines, a diagnostic or prognostic model,  explanations, and a web-based interface for clinicians to interact with and use the derived model.}\label{fig:adjutorium}
\end{figure*}

\section{{\proposedf}}
\label{sec:methods}

{\proposedf} is an algorithmic framework and software package that allows healthcare professionals to leverage ML to develop diagnostic and prognostic models.
Our framework employs automated machine learning \cite{Feurer2015} to tackle the challenges faced by clinical users.
By automating the optimization of ML pipelines involving data processing, model development, and model training, we reduce the burden on technical experts and turn deriving ML models from an art to a science, democratizing machine learning and opening the field to non-ML domain experts, such as clinicians.
We believe that {\proposedf} represents a step-change in algorithmic and software capabilities and can unlock the potential of ML in healthcare for clinical researchers \textit{without} the requirement for extensive technical capabilities.

{\proposedf} empowers healthcare professionals with the following capabilities:
\begin{enumerate}
    \item Build highly performant ML pipelines for classification, regression and time-to-event analysis, optimized specifically for the data at hand.
    \item Understand when ML provides benefits over traditional regression models, and thus when ML is valuable.
    \item Enable principled selection of variables and allow users to understand the value of information.
    \item Explain and debug ML models using diverse interpretability methods.
    \item Update systems whenever the available data changes to ensure the best possible clinical models. 
    \item Provide confidence in the reproducibility of models. 
\end{enumerate}

\subsection*{Overview}

After a clinician has determined an appropriate cohort of patients and an outcome of interest, the {\proposed} framework handles all steps in the computational pipeline: missing data imputation, feature processing, model selection and fitting, model interpretability or explanations, and production of clinical demonstrators.
Together, we believe {\proposed} significantly reduces the technical expertise necessary to derive powerful prognostic models, empowering clinical users and democratizing machine learning in healthcare.
An overview of {\proposedf} is provided in Figure \ref{fig:adjutorium}. Below, we provide a summary of each of the core components of {\proposed}. 
%Additional details can be found in Supplemental XXX.

\subsection*{Missing data imputation}

Medical datasets are often incomplete; however, most models require complete data as input, thus imputation is a necessary first step. 
There are many different imputation methods available, ranging from traditional statistical approaches such as mean imputation to well known alternatives such as MICE \cite{mice2011} and MissForest \cite{stekhoven2011missforest}. We include eight common imputation algorithms in {\proposed} for users to select if they desire a specific imputation method.

In addition, we also include a state-of-the-art AutoML approach for imputation, HyperImpute \cite{Jarrett2022}. HyperImpute is a generalized iterative imputation algorithm that automatically configures feature-wise imputation models. HyperImpute inherits the usual properties of classical iterative imputation algorithms \cite{mice2011,Liu2013,van2018flexible} while benefiting from an automated model selection and hyperparameter optimization procedure that allows the most appropriate model to be chosen for each feature. HyperImpute optimizes over five classes of model, with a total of 29 configurable hyperparameters.
For additional details, we refer to the recent technical report detailing HyperImpute \cite{Jarrett2022}. 
HyperImpute is the recommended imputation strategy in {\proposed}, unless a specific method is preferred by the user. Alternatively, the imputation step can be jointly optimized as part of a larger pipeline. 

\subsection*{Developing optimized ML pipelines}

After imputation, we construct ML pipelines consisting of feature processing, model selection, and model fitting. 
Given an objective function, these steps are jointly optimized using AutoML. There are several possible choices for the pipeline search algorithm, such as Bayesian optimization \cite{Alaa2018,wang2017batched} or bandit-based approaches \cite{Lisha2018hyperband}. 
%These steps are jointly optimized using an AutoML algorithm based on Hyperband \cite{Lisha2018hyperband}, a state-of-the-art bandit-based approach that has been demonstrated to outperform Bayesian optimization approaches, such as the method used in the first version of AutoPrognosis \cite{Alaa2018}. 
A key difference in this work is the extension of such approaches beyond hyperparameter optimization, the typical use of AutoML, to accommodate more general configuration spaces that encompass ML pipelines.
{\proposed} is flexible to the choice of AutoML search algorithm and can be extended as new approaches are developed. Currently, our default approach is based on Bayesian optimization.
In Table \ref{tbl:algorithms}, we provide a list of the algorithms currently implemented in {\proposedf}, together with the number of hyperparameters optimized over for each method. We emphasize the extendability of our approach to new methods, algorithms, and hyperparameters. 

\begin{table}[]
\resizebox{\textwidth}{!}{
\begin{tabular}{l|lllll}
\toprule
\textbf{Pipeline Stage}           & \multicolumn{5}{c}{\textbf{Algorithm (No. Hyperparameters Optimized by {\proposed})}}                                                                          \\
\midrule
\midrule
\textbf{Imputation}               & HyperImpute                    & Mean (0)                & Median (0)                        & Most-Frequent (0)           & MissForest (2)       \\
                                  & (M)ICE (0)                     & SoftImpute (2)          & EM (1)                            & Sinkhorn (6)                & None (0)             \\
\midrule                                  
\textbf{Dimensionality}           & Fast ICA (1)                   & Feat. Agg. (1)          & Gauss. Rand. Proj. (1)            & PCA (1)                     & Var. Thresh. (0)     \\
\textbf{Reduction}                &                                &                         &                                   &                             &                      \\
\midrule
\textbf{Feature}                  & L2 Norm. (0)                   & Max (0)                 & MinMax (0)                        & Normal Trans. (0)           & Quant. Trans. (0)    \\
\textbf{Scaling}                  & Unif. Trans. (0)               & None (0)                &                                   &                             &                      \\
\midrule
\textbf{Classification}           & ADABoost (3)                   & Bagging (4)             & Bernoulli NB (1)                  & CatBoost (2)                & Decision Tree (1)    \\
                                  & ExtraTree (1)                  & Gauss. NB (0)           & Grad. Boost. (3)                  & Hist. Grad. Boost. (2)      & KNN (4)              \\
                                  & LDA (0)                        & Light GBM (6)           & Linear SVM (1)                    & Log. Reg. (4)               & Multi. NB (1)        \\
                                  & Neural Net. (6)                & Perceptron (2)          & QDA (0)                           & Random Forest (5)           & Ridge Class. (1)     \\
                                  & TabNet (8)                     & XGBoost (11)            &                                   &                             &                      \\
\midrule
\textbf{Regression}               & Bayesian RR (1)                & CatBoost (2)            & Linear (0)                        & MLP (0)                     & Neural Net. (6)      \\
                                  & TabNet (8)                     & XGBoost (2)             &                                   &                             &                      \\
\midrule
\textbf{Survival}                 & Cox PH (2)                     & CoxNet (6)              & DeepHit (7)                       & LogLogistic AFT (1)         & LogNorm. AFT (2)     \\
\textbf{Analysis}                 & Surv. XGB (4)                  & Weibull AFT (2)         &                                   &                             &                      \\
\midrule
\textbf{Interpretability}         & INVASE                         & KernelSHAP              & LIME                              & Effect Size                 & Shap Permutation     \\
                                  & SimplEx                        & Symb. Persuit           &                                   &                             &                      \\
\bottomrule
\end{tabular}
}
\caption{List of algorithms currently included in {\proposedf}, grouped by pipeline stage. Numbers in brackets correspond to the number of hyperparameters optimized over by {\proposed}. {\proposed} is readily extendable to additional methods, algorithms, and hyperparameters.}
\label{tbl:algorithms}
\end{table}

\textit{Feature processing.} While imputation ensures data is complete, preprocessing datasets is a common requirement for many ML estimators. In particular, feature scaling to normalize the range or the shape of features can significantly affect performance \cite{Crone2006}.
{\proposed} can optimize over five dimensionality reduction and six feature scaling algorithms.

\textit{Model selection and fitting.} Next, a model and hyperparameters must be selected. This is a key step as suboptimal choice of model or hyperparameters can significantly affect the performance of the resulting ML system.
{\proposed} contains 22 classification algorithms, seven regression algorithms, and seven methods for survival analysis. Together with a range of hyperparameters, this defines a broad algorithmic search space. While navigating this space manually by hand is extremely challenging, {\proposed} learns relationships between different settings to efficiently arrive at an optimized solution. 
Finally, {\proposed} combines the best performing models into a single ensemble using the posterior belief of the AutoML algorithm.

\subsection*{Model explanations}

Predictive models alone are not sufficient and more must be done to engender model trust from both clinical users \cite{Rajpurkar2022} and regulatory bodies \cite{FDA2019,Mourby2021,Kattan2016}. 
Consequently, {\proposed} contains a suite of methods for explaining ML models. We have included feature-based interpretability methods, such as SHAP \cite{lundberg2017unified}, that allow us to understand the importance of individual features, as well as an example-based interpretability method, SimplEx \cite{Crabbe2021Simplex}, that explains the model output for a particular sample with examples of similar instances, similar to case-based reasoning.  
Furthermore, sometimes outputs of a specific form are required, such as explicit risk equations \cite{Kattan2016}. We have therefore included the ability to convert optimized models into transparent risk equations using symbolic regression \cite{Crabbe2020}. 

\subsection*{Demonstrators}

In order for risk scores to be useful, they need to be readily available to clinical practitioners. To facilitate this, {\proposed} allows interactive demonstrators to be produced for clinical use. 
We build our clinical demonstrators on top of the open-source Streamlit package \cite{streamlit}.
Compared to traditional solutions, these require almost no technical capabilities to set up, and the standardized nature simplifies adoption for end-users. 

\section{Illustrative application of {\proposedf}}
\label{sec:results}

In this section, we show how {\proposedf} can be applied to address the challenges described in Section \ref{sec:challenges}.
We demonstrate the application of {\proposedf} using an illustrative scenario: prognostic risk prediction of developing diabetes using a cohort of 502,467 individuals from UK Biobank.
Our goal is \textit{not} to develop the best model for diabetes risk prediction possible, but instead to exemplify how our tool can be used.

In our use-scenario, we show that the model derived with {\proposed} outperforms risk models currently used in clinical practice and quantify the benefit of ML methods over Cox proportional hazard models.
In addition, we show how the model interpretability components of {\proposed} can be used to understand the drivers of predictions and identify novel risk factors not incorporated into previous risk scores.  
Finally, we use {\proposed} to share the diabetes risk score as a web-based decision support tool which can be publicly accessed by patients and clinicians worldwide.\footnote{\url{https://autoprognosis-biobank-diabetes.streamlitapp.com/}}

While we illustrate risk prediction of developing diabetes using a cohort from UK Biobank, {\proposed} can be applied to construct diagnostic and prognostic models for any disease or clinical outcome. Furthermore, {\proposed} is applicable to classification and regression tasks, in addition to survival analysis.

\subsection{Designing experiments}

\textbf{Selecting which dataset to use}
{\proposed} can be used with data from many different origins, such as biobanks \cite{Alaa2019}, registries \cite{Alaa2018CF,Alaa2021}, and private hospital data \cite{Shah2021THA}.
Here, we use the UK Biobank, due to its availability and popularity as a resource for healthcare researchers. UK Biobank enrolled half a million participants from 22 assessment centers across England, Wales, and Scotland between 2006 and 2010 \cite{Sudlow2015}, with follow-up data collected from hospital records \cite{Adamska2015}.
From UK Biobank, we extracted a cohort of participants who were 40 years of age or older with no diagnosis or history of diabetes at baseline; the primary outcome was diagnosis of diabetes within a 10 year horizon. We selected diabetes as our outcome of interest due to its global prevalence and role as a risk factor for a multitude of other indications \cite{world2016global}.

\textbf{Selecting variables}
Variables can be selected for inclusion in a study in a myriad of ways. Often, healthcare professionals will select a subset of exploratory features that are of particular interest to them. This could be due to supporting medical literature, to explore a hypothesis, or based on features included in existing risk scores. Alternatively, we can always chose to initially include all available variables.
Here, we selected an initial set of 109 exploratory features based on their general clinical availability, discussions with clinicians, and features used by existing risk scores.
We purposefully selected almost an order of magnitude increase compared to existing risk scores to illustrate how {\proposed} can be used in such a scenario. 

\textbf{Selecting benchmarks}
Often, existing risk scores will exist for the outcome of interest; this is certainly true for diabetes, where several risk scores that estimate the probability of developing diabetes are currently used in clinical practice. Therefore, we use the following as baseline risk scores:
\begin{itemize}
\item \textbf{ADA:} The American Diabetes Association (ADA) risk score \cite{Bang2009} is a points-based score employing six features, namely age, sex, family history of diabetes, history of hypertension, obesity, and physical activity. 
\item \textbf{FINRISK:} A risk score for diabetes was derived from FINRISK, a large population survey in Finland, based on age, body mass index (BMI), waist circumference, history of antihypertensive drug treatment and high blood glucose, physical activity, and daily consumption of fruits, berries, or vegetables \cite{Lindstrom2003}. 
\item \textbf{DiabetesUK:} The risk score from Diabetes UK uses seven features: gender, age, ethnicity, family history, waist size, BMI, and high blood pressure requiring treatment.
\item \textbf{QDiabetes:} Finally, QDiabetes \cite{Hippisley-Cox2017Diabetes} consists of three separate models depending on the clinical information available and stage of risk screening. Model A uses 16 non-laboratory features that do not require a blood test and is intended primarily as an initial screening tool. 
%These are age, ethnicity, deprivation, BMI, smoking, family history of diabetes, cardiovascular disease, treated hypertension, regular use of corticosteroids, atypical antipsychotics, statins, schizophrenia or bipolar affective disorder, learning disability, and gestational diabetes and polycystic ovary syndrome in women. 
Models B and C include the same variables as Model A together with fasting blood glucose and hemoglobin A1c (HbA1c), respectively, with the aim of refining risk assessment following a blood test. 
\end{itemize}

In addition to the baseline risk scores, a comparison with traditional modeling approaches can be made using {\proposed}. We demonstrate this by fitting Cox proportional hazard (Cox PH) \cite{Cox1972} models using the same features as each of the baseline risk scores. These models can be thought of as variants of the respective risk scores calibrated to the specific dataset. 

\begin{table*}[ht]
	\vskip -0.05in
	\caption{Diabetes risk prediction results. The risk scores automatically derived by {\proposed} outperform the existing risk scores and Cox PH models retrained on the same features. Mean performance reported with 95\% confidence interval.} \label{tbl:diabetes}
	\vskip 0.15in
	\begin{center}
    \begin{small}
    \begin{sc}
    \resizebox{\textwidth}{!}{
		\begin{tabular}[b]{c c c c}
			\toprule
			\textbf{Method}         & \textbf{C-index $\uparrow$} & \textbf{Brier score $\downarrow$} & \textbf{AUROC $\uparrow$} \\
			\midrule
            ADA                     & 0.696 $\pm$ 0.015 & 0.011 $\pm$ 0.000 & 0.697 $\pm$ 0.018 \\
            FINRISK                 & 0.728 $\pm$ 0.029 & 0.019 $\pm$ 0.000 & 0.729 $\pm$ 0.020 \\
            DiabetesUK              & 0.759 $\pm$ 0.013 & 0.016 $\pm$ 0.000 & 0.759 $\pm$ 0.019 \\
            QDiabetes Model A       & 0.794 $\pm$ 0.022 & 0.008 $\pm$ 0.000 & 0.795 $\pm$ 0.017 \\
            QDiabetes Model B       & 0.788 $\pm$ 0.019 & 0.015 $\pm$ 0.000 & 0.788 $\pm$ 0.013 \\
            QDiabetes Model C       & 0.839 $\pm$ 0.021 & 0.005 $\pm$ 0.000 & 0.840 $\pm$ 0.010 \\
            \midrule
            Cox PH (ADA)            & 0.774 $\pm$ 0.027 & 0.002 $\pm$ 0.000 & 0.774 $\pm$ 0.020 \\
            Cox PH (FINRISK)        & 0.786 $\pm$ 0.023 & 0.002 $\pm$ 0.000 & 0.786 $\pm$ 0.026 \\
            Cox PH (DiabetesUK)     & 0.794 $\pm$ 0.023 & 0.002 $\pm$ 0.000 & 0.794 $\pm$ 0.022 \\
            Cox PH (QDiabetes C)    & 0.858 $\pm$ 0.007 & 0.002 $\pm$ 0.000 & 0.860 $\pm$ 0.018 \\
			\midrule
            \textbf{{\proposedf}}   & \textbf{0.888 $\pm$ 0.007} & \textbf{0.002 $\pm$ 0.000} & \textbf{0.888 $\pm$ 0.012} \\
            {\proposed} (19 feat.)  & 0.870 $\pm$ 0.011 & 0.002 $\pm$ 0.000 & 0.867 $\pm$ 0.020 \\
			\bottomrule
        \end{tabular} 
        }
 	\end{sc}
 	\end{small}
 	\end{center}
    \hfill
    \vskip -0.3in
\end{table*}

\subsection{Using {\proposedf} to address the challenges of diagnostic and prognostic modeling}

Through the lens of our example (diabetes risk prediction), we demonstrate how {\proposedf} can be used to address the challenges of diagnostic and prognostic modeling introduced in Section \ref{sec:challenges}.

\vspace{\baselineskip}
\textbf{Challenge 1. Developing powerful ML pipelines}

We begin by using {\proposed} to derive a clinical risk score for diabetes.
We evaluate the performance of the models using concordance index (C-index) to assess model discrimination, Brier score to assess calibration, and the area under the receiver-operating curve (AUROC) to assess prediction accuracy. 
We perform imputation fives times and then conduct 3-fold cross validation for each of the imputed datasets.
As seen in Table \ref{tbl:diabetes}, the risk score developed by {\proposed} significantly outperforms all baseline risk scores and Cox PH models (p-value $<0.001$), achieving a C-index on the validation cohort of 0.888 (95\% confidence interval: 0.881-0.895).
This compares to 0.696 (0.681-0.711) for the ADA score, 0.728 (0.699-0.757) for FINRISK, 0.759 (0.746-0.772) for DiabetesUK, and 0.839 (0.818-0.860) for the best performing QDiabetes model (Model C). 
Cox PH models fit with the same risk factors as the clinical risk scores achieved improved performance (C-indices: 0.774, 0.786, 0.794, and 0.858, respectively), but exhibit lower performance than {\proposed}.

\begin{figure*}[t]%
\centering
\vspace{-0.1in}
\includegraphics[trim={0 0cm 0 0cm}, clip, width=0.7\textwidth]{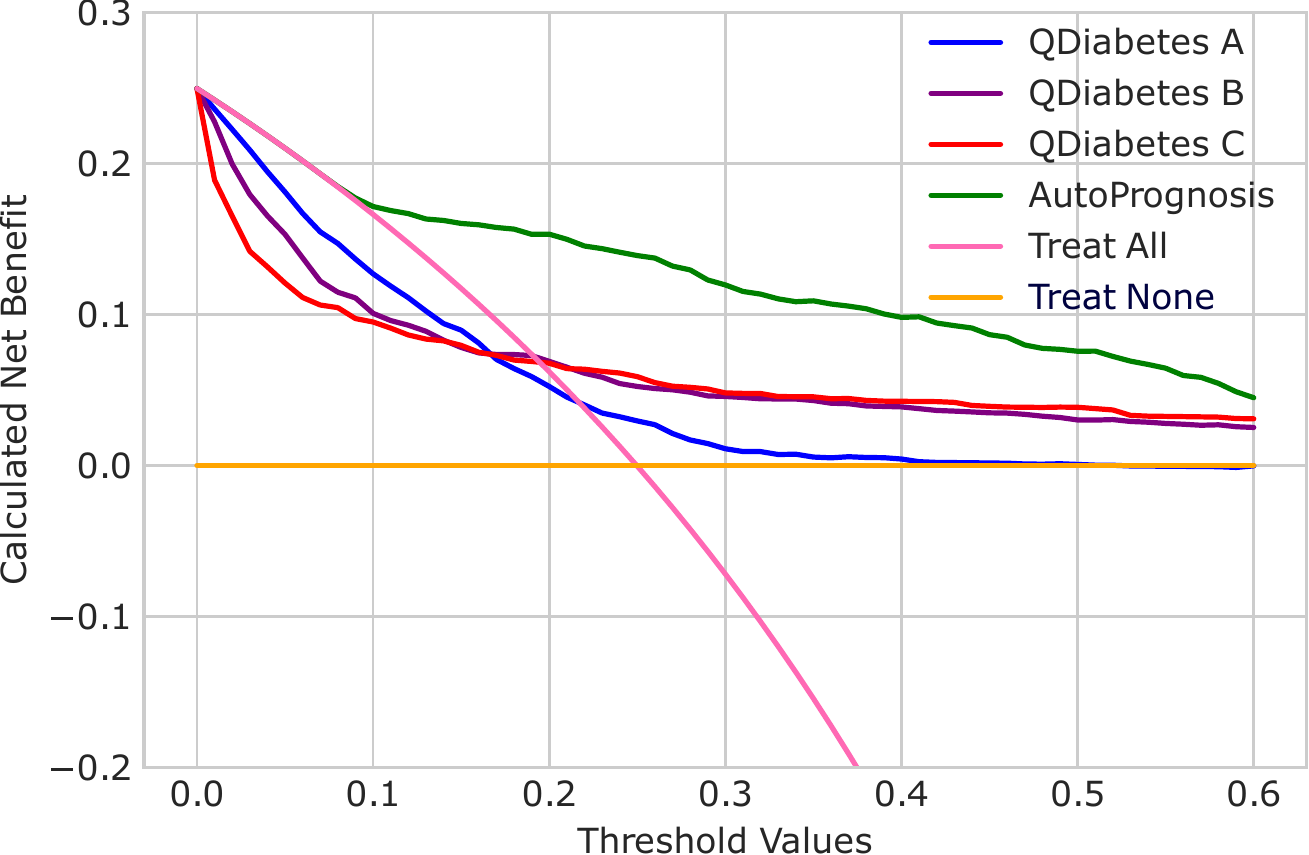}
\vspace{-0.1in}
\caption{Decision curve analysis. {\proposed} exhibits higher net benefit at all decision thresholds compared to existing risk scores and baseline treatment plans. }\label{fig:diabetes-dca}
\end{figure*}

As an alternate way of understanding the clinical impact of our results, we performed decision curve analysis and calculated the clinical net benefit across a range of risk threshold probabilities.
We compared the predicted risk by 
{\proposed} with the QDiabetes models, the best performing of the existing clinical risk scores, as well as baseline strategies to treat all patients (Treat All) or no-one (Treat None).
Decision curve analysis further demonstrates the benefit of {\proposed} compared to existing risk scores for diabetes. At all decision thresholds, {\proposed} offers greater net benefit and is the only score to outperform ``Treat All'' between the 0.1 and 0.2 thresholds, and the only model to perform in-line with ``Treat All'' below a threshold of 0.1.

\vspace{\baselineskip}

\begin{wraptable}{r}{0.45\textwidth}
%\begin{table*}[ht]
	\vskip -0.35in
	%\caption{Quantifying the value of ML. Using both QDiabetes Model C features and our expanded feature set, the risk scores automatically derived by {\proposed} significantly outperform Cox PH models (p-values: XXX and 0.005, respectively).} \label{tbl:diabetes_value_ml}
 \caption{Quantifying the value of ML. The risk score automatically derived by {\proposed} significantly outperforms a Cox PH model trained on the same features (p-value: 0.005).} \label{tbl:diabetes_value_ml}
	\vskip 0.0in
	\begin{center}
    \begin{small}
    \begin{sc}
    \resizebox{0.42\textwidth}{!}{
		\begin{tabular}[b]{c c}
			\toprule
			\textbf{Method}         & \textbf{C-index $\uparrow$}    \\
			%\midrule
            %\multicolumn{2}{c}{\textit{QDiabetes Model C Variables}} \\
            %Cox PH                  & 0.858 $\pm$ 0.007              \\
            %{\proposed}             &                                \\
			\midrule
            \multicolumn{2}{c}{\textit{All Variables}}               \\
            Cox PH                  & 0.883 $\pm$ 0.010              \\
            {\proposed}             & 0.888 $\pm$ 0.007              \\
			\bottomrule
        \end{tabular} 
        }
 	\end{sc}
 	\end{small}
 	\end{center}
    \hfill
    \vskip -0.3in
%\end{table*}
\end{wraptable}
\textbf{Challenge 2. Understanding when ML is necessary and its value}
Table \ref{tbl:diabetes} demonstrates the benefit of {\proposed} compared to existing risk scores and Cox PH models retrained on the same features. We now directly compare {\proposed} to Cox PH models on the same training data to understand if ML is needed for this problem. In Table \ref{tbl:diabetes_value_ml}, we show the performance of {\proposed} and a Cox PH model using the full feature set considered. We see that while some of the benefit is due to the additional features, there remains value in the improved modeling approach, even for identical feature sets (p-value: 0.005).   

\vspace{\baselineskip}
\textbf{Challenge 3. Determining the value of information}

\begin{wrapfigure}{r}{0.5\textwidth}
%\begin{figure*}[ht]%
\vskip -0.2in
\centering
\includegraphics[width=0.45\textwidth]{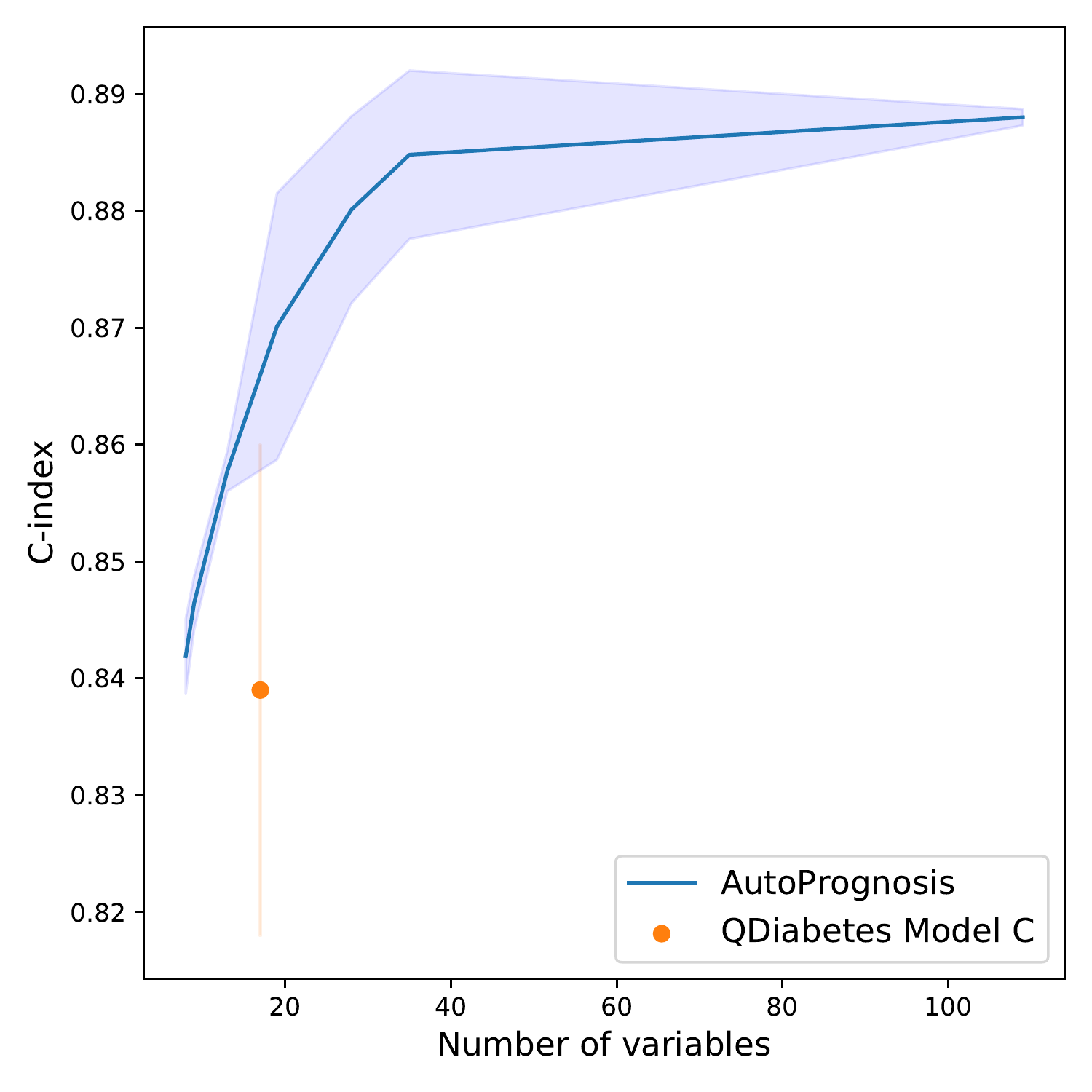}
\caption{Value of information. We evaluate {\proposed} using different numbers of features, corresponding to different effect size thresholds. The feature efficiency is compared to QDiabetes Model C, the best performing existing risk score. }\label{fig:perf_n_feat}
\vskip -0.1in
%\end{figure*}
\end{wrapfigure}
Understanding the predictive power of variables is key and often there is a trade-off (e.g. cost or time) in clinical practice to acquiring additional variables. 
We evaluate {\proposed} using different subsets of features. We selected features using the magnitude of the effect size. We measure the distributional shift for an increase in predicted risk using Cohen's D \cite{cohen2013statistical}, and select features with effect sizes exceeding the thresholds \{0.5, 0.6, 0.7, 0.8, 0.9, 1.0\}.
Even using only eight features, {\proposed} slightly outperforms the best performing existing risk score, QDiabetes Model C (Figure \ref{fig:perf_n_feat}). 
As the number of features increases, performance rapidly increases until 35 features are used (effect size: 0.5). After this point, while there is some gain from additional features, it could be considered marginal given the number of additional features employed.

\vspace{\baselineskip}
\textbf{Challenge 4. Understanding and debugging ML models}

\begin{wrapfigure}{r}{0.5\textwidth}
%\begin{figure*}[ht]%
\vskip -0.2in
\centering
\includegraphics[width=0.49\textwidth]{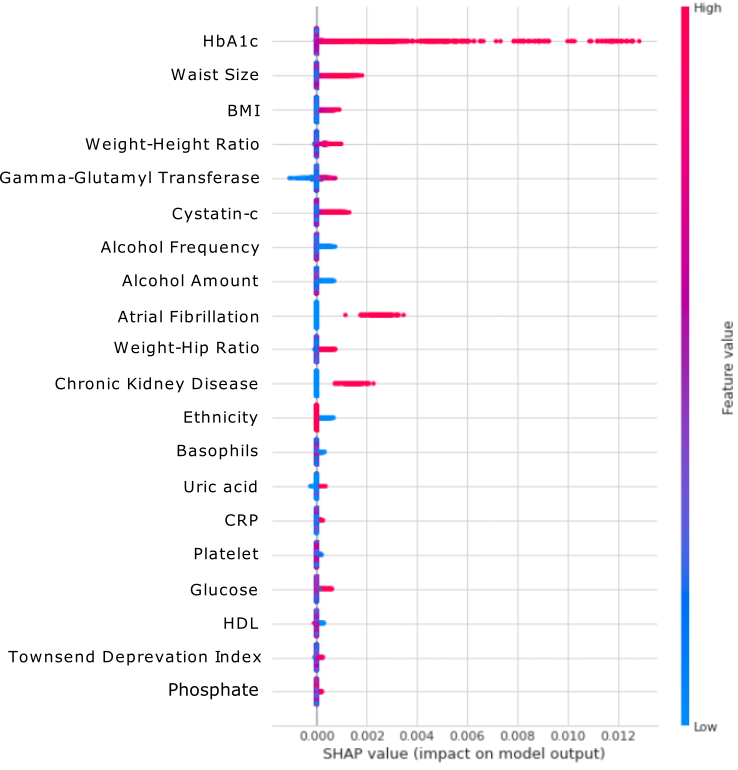}
\caption{SHAP values for the most important features. }\label{fig:diabetes_ap_shap_values}
\vskip -0.1in
%\end{figure*}
\end{wrapfigure}

Highly predictive models alone are insufficient and it is necessary to understanding which features are important. 
We demonstrate how the interpretability methods incorporated in {\proposedf} can be used to understand how ML models make predictions and debug their behavior. We begin by examining the SHAP values \cite{lundberg2017unified} to explain the key contributors to model performance. Figure \ref{fig:diabetes_ap_shap_values} shows the top 20 features. Encouragingly, these features are largely consistent with clinical knowledge, providing evidence that the model is acting in a desirable manner. 
Several of the top risk factors, such as HbA1c, waist size, and body mass index, were also included in previous risk scores. However, a number of additional features, including both laboratory and non-laboratory tests, were deemed important.
A number of these features have been shown to be risk factors for diabetes (e.g. gamma-glutamyl transferase \cite{Nano2017ggt}), but have not been incorporated into other risk scores.
Of the existing risk factors, we find that HbA1c is significantly more important to the predictions of {\proposed} than blood glucose, which is consistent with our earlier experiments that showed QDiabetes Model C (which uses HbA1c) outperforms Model B (which uses blood glucose) on the UK Biobank population. 

Finally, several features commonly incorporated in previous risk scores are notably missing: for example age and sex. One explanation could be that UK Biobank contains a limited age range (40-69 at enrollment), and thus the role of age could be reduced over that range. However, increasingly, younger individuals are being diagnosed with diabetes \cite{IDF2013}, which could also explain the omission of age as a key risk factor. In the case of sex, while it was once assumed that there were sex differences, diabetes is equally prevalent among men and women in most populations \cite{Gale2001}. 

To illustrate debugging, we consider the development of diabetes in individuals with differing HbA1c levels. We divide the overall cohort into two approximately equal parts using the median HbA1c value of 4.69\%. This equates to splitting the population into a low-normal subgroup and a high-normal and elevated subgroup \cite{ADA_website}. 
%The normal range for HbA1c for individuals without diabetes is between 4\% and 5.6\%, HbA1c levels between 5.7\% and 6.4\% is considered prediabetic, while patients with levels of 6.5\% or above are considered to be diabetic \cite{ADA_website}.

\begin{table*}[t]
	\vskip -0.05in
	\caption{Performance of diabetes risk scores for subgroups defined by HbA1c.} \label{tbl:diabetes_hba1c}
	\vskip 0.15in
	\begin{center}
    \begin{small}
    \begin{sc}
    \resizebox{0.95\textwidth}{!}{
		\begin{tabular}[b]{c c c c c}
			\toprule
			\textbf{Method}    
                %& \multicolumn{2}{c}{\textbf{HbA1c $\pmb{<4.69\%}$}} & \multicolumn{2}{c}{\textbf{HbA1c $\pmb{\geq 4.69\%}$}} \\
                %& \textbf{C-index} & \textbf{AUROC} & \textbf{C-index} & \textbf{AUROC} \\
			%\midrule
            %QDiabetes (Model A)     & 0.771 $\pm$ 0.053 & 0.772 $\pm$ 0.009 & 0.775 $\pm$ 0.016 & 0.775 $\pm$ 0.023 \\
            %QDiabetes (Model B)     & 0.738 $\pm$ 0.031 & 0.738 $\pm$ 0.007 & 0.773 $\pm$ 0.010 & 0.773 $\pm$ 0.017 \\
            %QDiabetes (Model C)     & 0.735 $\pm$ 0.052 & 0.736 $\pm$ 0.022 & 0.855 $\pm$ 0.008 & 0.856 $\pm$ 0.004 \\
            %\midrule
            %{\proposed}             & 0.818 $\pm$ 0.047 & 0.807 $\pm$ 0.013 & 0.889 $\pm$ 0.011 & 0.896 $\pm$ 0.009 \\
                & \multicolumn{2}{c}{\textbf{C-index}} & \multicolumn{2}{c}{\textbf{AUROC}} \\
                & \textbf{HbA1c $\pmb{<4.69\%}$} & \textbf{HbA1c $\pmb{\geq 4.69\%}$} & \textbf{HbA1c $\pmb{<4.69\%}$} & \textbf{HbA1c $\pmb{\geq 4.69\%}$} \\
			\midrule
            QDiabetes Model A     & 0.771 $\pm$ 0.053 & 0.775 $\pm$ 0.016 & 0.772 $\pm$ 0.009 & 0.775 $\pm$ 0.023 \\
            QDiabetes Model B     & 0.738 $\pm$ 0.031 & 0.773 $\pm$ 0.010 & 0.738 $\pm$ 0.007 & 0.773 $\pm$ 0.017 \\
            QDiabetes Model C     & 0.735 $\pm$ 0.052 & 0.855 $\pm$ 0.008 & 0.736 $\pm$ 0.022 & 0.856 $\pm$ 0.004 \\
            \midrule
            {\proposedf}          & 0.818 $\pm$ 0.047 & 0.889 $\pm$ 0.011 & 0.807 $\pm$ 0.013 & 0.896 $\pm$ 0.009 \\
			\bottomrule
        \end{tabular} 
        }
 	\end{sc}
 	\end{small}
 	\end{center}
    \hfill
    \vskip -0.1in
\end{table*}

We evaluated {\proposed} and the QDiabetes models on these two cohorts (Table \ref{tbl:diabetes_hba1c}).
Despite displaying better performance across the entire dataset, QDiabetes Model C \textit{under}performs Model A for patients in the low-normal HbA1c cohort.
Conversely, {\proposed} performs best for both subgroups, although predicting future risk of diabetes is more challenging for low-normal HbA1c patients, in line with the other models.
This could suggest that QDiabetes Model C is overly reliant on HbA1c while {\proposed} has more accurately captured the risk factors for low HbA1c patients.

This raises the question of \textit{why} {\proposed} is able to issue more accurate predictions for the low-normal HbA1c cohort, in particular given HbA1c is ranked as the most important feature globally (Figure \ref{fig:diabetes_ap_shap_values}).
Table \ref{tbl:diabetes_hba1c_interpretability} shows the most important features (measured by risk effect size) for the two subgroups defined by HbA1c.
While there is significant overlap, there are five unique features in the top 20 for each cohort.
This type of analysis can help clinicians understand and debug the predictions of models not only for the entire population, but specific subgroups of interest.

\begin{table*}[ht]
	\vskip -0.05in
	\caption{The most important features for {\proposed} measured by risk effect size (value in parenthesis) for the two cohorts defined by median HbA1c. Features in \textcolor{blue}{blue} differ between the two cohorts.} \label{tbl:diabetes_hba1c_interpretability}
	\vskip 0.15in
	\begin{center}
    \begin{small}
    \begin{sc}
    \resizebox{0.8\textwidth}{!}{
		\begin{tabular}[b]{c c}
			\toprule
			\textbf{HbA1c $\pmb{<4.69\%}$} & \textbf{HbA1c $\pmb{\geq 4.69\%}$} \\
			\midrule
                \textcolor{blue}{Atrial fibrillation (3.0)} & \textcolor{blue}{HbA1c (3.0)} \\
                Waist Size (2.8)                            & \textcolor{blue}{Glucose (2.5)} \\
                Body Mass Index (2.7)                       & Weight/Height Ratio (1.5) \\
                Weight/Height Ratio (2.7)                   & Waist Size (1.5) \\
                Weight (2.7)                                & Body Mass Index (1.4) \\
                Hip Size (2.2)                              & Weight (1.3) \\
                Waist/Hip Ratio (1.8)                       & Waist/Hip Ratio (1.1) \\
                Cystatin-c (1.6)                            & Hip Size (1.1) \\
                \textcolor{blue}{Kidney Disease (1.5)}      & Alanine Transaminase (0.87) \\
                \textcolor{blue}{Uric Acid (1.3)}           & Triglycerides (0.76) \\
                Alanine Transaminase (1.1)                  & Gamma-Glutamyl Transferase (0.74) \\
                \textcolor{blue}{Anti-hypertensive Medication (1.1)} & \textcolor{blue}{HDL (0.71)} \\
                \textcolor{blue}{History of Hypertension (0.99)} & \textcolor{blue}{C-Reactive Protein (0.70)} \\
                Triglycerides (0.97)                        & Cystatin-c (0.68) \\
                Gamma-Glutamyl Transferase (0.96) & \textcolor{blue}{Sex Hormone-Binding Globulin (0.67)} \\
			\bottomrule
        \end{tabular} 
        }
 	\end{sc}
 	\end{small}
 	\end{center}
    \hfill
    \vskip -0.1in
\end{table*}

\begin{figure*}[ht]%
\centering
\includegraphics[width=\textwidth]{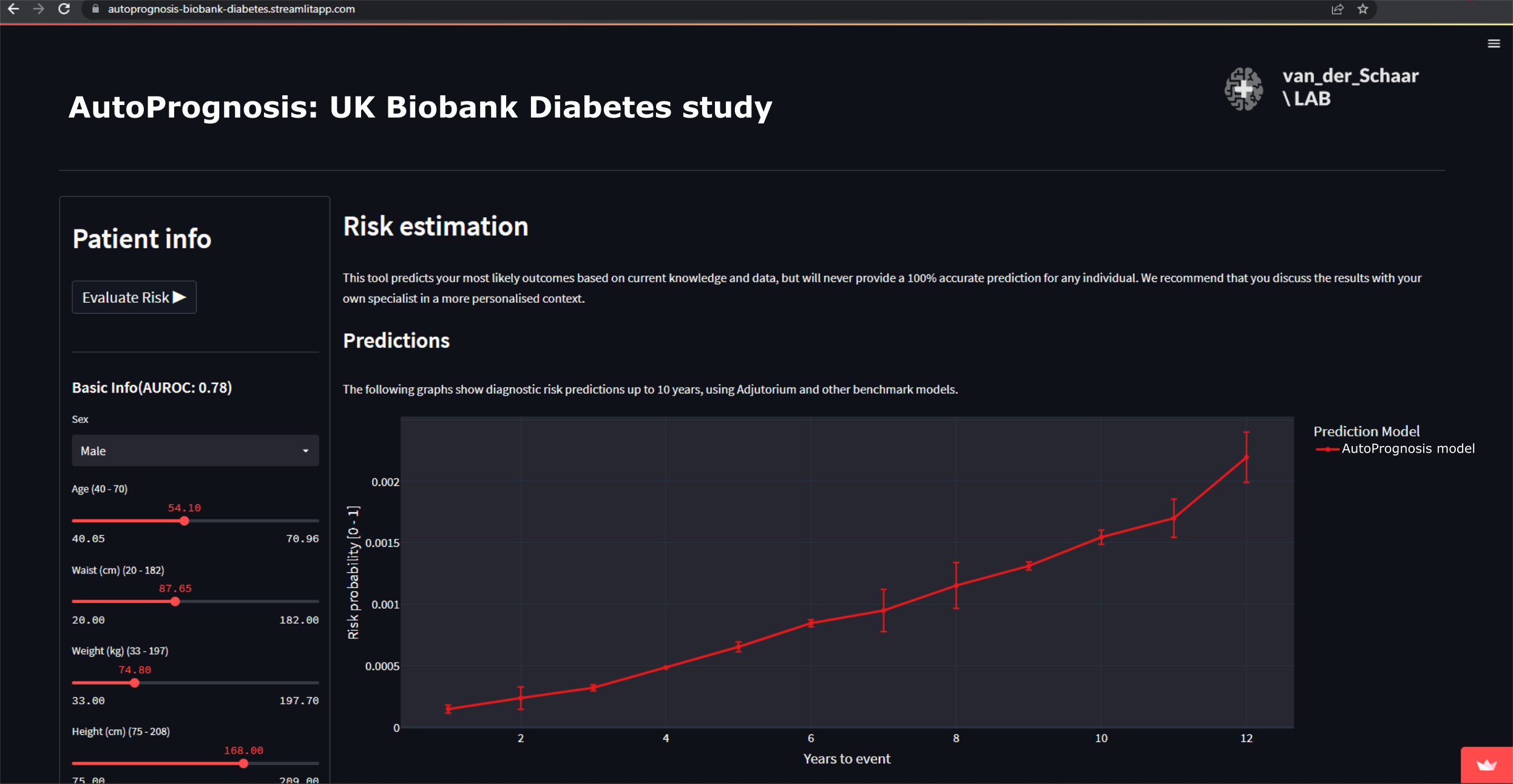}
\caption{Screenshot of an example clinical demonstrator produced by {\proposed}.}\label{fig:demonstrator}
\end{figure*}

\vspace{\baselineskip}
\textbf{Challenge 5. Making ML models accessible and usable}

Finally, we end our illustrative scenario with an example web-based demonstrator enabling the use of the risk model derived by {\proposed}. The web application can be accessed at \url{https://autoprognosis-biobank-diabetes.streamlitapp.com/}. A screenshot is provided in Figure \ref{fig:demonstrator}. 

\section{Using {\proposed} in Healthcare and Beyond} \label{sec:discussion}

Advances in ML algorithms harbor the potential to transform healthcare; however, major challenges continue to limit their adoption in medicine.
In this work, we define these challenges and describe the first integrated, automated framework for diagnostic and prognostic modeling, {\proposedf}, that is designed explicitly to overcome each obstacle in a way that is accessible to non-expert users, democratizing model construction, understanding, debugging, and sharing.

While we have provided an illustrative example of how {\proposed} can be used, the key finding reported here is \textit{not} the performance of a single illustrative model, but rather the way in which it was built. 
We believe {\proposedf} is a necessary development in the journey towards widespread adoption of ML systems in clinical practice and hope that researchers will engage with this tool. 
Rather than marginalizing healthcare experts, we believe {\proposed} places them at the center and empowers them to create new clinical tools.
As part of this journey, we will continue to add new features and improve {\proposed}.
Finally, while the focus and motivation for {\proposed} is medicine, it has not escaped our notice that {\proposed} can be used to construct predictive models and risk scores for applications beyond healthcare.

\appendix

%\section{Sample Appendix Section}
%tempor section \ref{sec:intro} incididunt ut labore et dolore 

 \bibliographystyle{elsarticle-num} 
 %\bibliography{cas-refs}
 \bibliography{bibliography}

\begin{thebibliography}{10}
\expandafter\ifx\csname url\endcsname\relax
  \def\url#1{\texttt{#1}}\fi
\expandafter\ifx\csname urlprefix\endcsname\relax\def\urlprefix{URL }\fi
\expandafter\ifx\csname href\endcsname\relax
  \def\href#1#2{#2} \def\path#1{#1}\fi

\bibitem{Topol2019}
E.~J. Topol, High-performance medicine: {T}he convergence of human and
  artificial intelligence, Nature Medicine 25~(1) (2019) 44--56.
\newblock \href {https://doi.org/10.1038/s41591-018-0300-7}
  {\path{doi:10.1038/s41591-018-0300-7}}.

\bibitem{Gerke2020}
S.~Gerke, T.~Minssen, G.~Cohen, Chapter 12 - {E}thical and legal challenges of
  artificial intelligence-driven healthcare, in: A.~Bohr, K.~Memarzadeh (Eds.),
  Artificial Intelligence in Healthcare, Academic Press, 2020, pp. 295--336.
\newblock \href
  {https://doi.org/https://doi.org/10.1016/B978-0-12-818438-7.00012-5}
  {\path{doi:https://doi.org/10.1016/B978-0-12-818438-7.00012-5}}.

\bibitem{Sun2019}
T.~Q. Sun, R.~Medaglia, Mapping the challenges of artificial intelligence in
  the public sector: {E}vidence from public healthcare, Government Information
  Quarterly 36~(2) (2019) 368--383.
\newblock \href {https://doi.org/https://doi.org/10.1016/j.giq.2018.09.008}
  {\path{doi:https://doi.org/10.1016/j.giq.2018.09.008}}.

\bibitem{Yu2018}
K.-H. Yu, A.~L. Beam, I.~S. Kohane, Artificial intelligence in healthcare,
  Nature Biomedical Engineering 2~(10) (2018) 719--731.
\newblock \href {https://doi.org/10.1038/s41551-018-0305-z}
  {\path{doi:10.1038/s41551-018-0305-z}}.

\bibitem{Rajpurkar2022}
P.~Rajpurkar, O.~Chen, Emmaand~Banerjee, E.~J. Topol, {AI} in health and
  medicine, Nature Medicine 28~(1) (2022) 31--38.
\newblock \href {https://doi.org/10.1038/s41591-021-01614-0}
  {\path{doi:10.1038/s41591-021-01614-0}}.

\bibitem{Petersson2022}
L.~Petersson, I.~Larsson, J.~M. Nygren, P.~Nilsen, M.~Neher, J.~E. Reed,
  D.~Tyskbo, P.~Svedberg, Challenges to implementing artificial intelligence in
  healthcare: {A} qualitative interview study with healthcare leaders in
  {S}weden, BMC Health Services Research 22~(1) (2022) 850.
\newblock \href {https://doi.org/10.1186/s12913-022-08215-8}
  {\path{doi:10.1186/s12913-022-08215-8}}.

\bibitem{Alaa2018}
A.~Alaa, M.~van~der Schaar, {A}uto{P}rognosis: {A}utomated clinical prognostic
  modeling via {B}ayesian optimization with structured kernel learning, In:
  Proceedings of the 35th International Conference on Machine Learning 80
  (2018) 139--148.

\bibitem{Alaa2019}
A.~M. Alaa, T.~Bolton, E.~Di~Angelantonio, J.~H.~F. Rudd, M.~van~der Schaar,
  Cardiovascular disease risk prediction using automated machine learning: {A}
  prospective study of 423,604 {UK Biobank} participants, PLOS ONE 14~(5)
  (2019) 1--17.
\newblock \href {https://doi.org/10.1371/journal.pone.0213653}
  {\path{doi:10.1371/journal.pone.0213653}}.

\bibitem{Alaa2018CF}
A.~M. Alaa, M.~van~der Schaar, Prognostication and risk factors for cystic
  fibrosis via automated machine learning, Scientific Reports 8~(1) (2018)
  11242.
\newblock \href {https://doi.org/10.1038/s41598-018-29523-2}
  {\path{doi:10.1038/s41598-018-29523-2}}.

\bibitem{Alaa2021}
A.~M. Alaa, D.~Gurdasani, A.~L. Harris, J.~Rashbass, M.~van~der Schaar, Machine
  learning to guide the use of adjuvant therapies for breast cancer, Nature
  Machine Intelligence 3~(8) (2021) 716--726.
\newblock \href {https://doi.org/10.1038/s42256-021-00353-8}
  {\path{doi:10.1038/s42256-021-00353-8}}.

\bibitem{Rahbar2020}
H.~Rahbar, D.~S. Hippe, A.~Alaa, S.~H. Cheeney, M.~van~der Schaar, S.~C.
  Partridge, C.~I. Lee, The value of patient and tumor factors in predicting
  preoperative breast {MRI} outcomes, Radiology: Imaging Cancer 2~(4) (2020)
  e190099.
\newblock \href {https://doi.org/10.1148/rycan.2020190099}
  {\path{doi:10.1148/rycan.2020190099}}.

\bibitem{Qian2021}
Z.~Qian, A.~M. Alaa, M.~van~der Schaar, {CPAS}: {T}he {UK}'s national machine
  learning-based hospital capacity planning system for {COVID-19}, Machine
  Learning 110~(1) (2021) 15--35.
\newblock \href {https://doi.org/10.1007/s10994-020-05921-4}
  {\path{doi:10.1007/s10994-020-05921-4}}.

\bibitem{Shah2021THA}
A.~A. Shah, S.~K. Devana, C.~Lee, R.~Kianian, M.~{van der Schaar}, N.~F.
  SooHoo, Development of a novel, potentially universal machine learning
  algorithm for prediction of complications after total hip arthroplasty, The
  Journal of Arthroplasty 36~(5) (2021) 1655--1662.e1.
\newblock \href {https://doi.org/https://doi.org/10.1016/j.arth.2020.12.040}
  {\path{doi:https://doi.org/10.1016/j.arth.2020.12.040}}.

\bibitem{Devana2021TKA}
S.~K. Devana, A.~A. Shah, C.~Lee, A.~R. Roney, M.~{van der Schaar}, N.~F.
  SooHoo, A novel, potentially universal machine learning algorithm to predict
  complications in total knee arthroplasty, Arthroplasty Today 10 (2021)
  135--143.
\newblock \href {https://doi.org/https://doi.org/10.1016/j.artd.2021.06.020}
  {\path{doi:https://doi.org/10.1016/j.artd.2021.06.020}}.

\bibitem{Shah2022spinalfusion}
A.~A. Shah, S.~K. Devana, C.~Lee, A.~Bugarin, E.~L. Lord, A.~N. Shamie, D.~Y.
  Park, M.~van~der Schaar, N.~F. SooHoo, Machine learning-driven identification
  of novel patient factors for prediction of major complications after
  posterior cervical spinal fusion, European Spine Journal 31~(8) (2022)
  1952--1959.
\newblock \href {https://doi.org/10.1007/s00586-021-06961-7}
  {\path{doi:10.1007/s00586-021-06961-7}}.

\bibitem{Shah2022C5}
A.~A. Shah, S.~K. Devana, C.~Lee, A.~Bugarin, M.~K. Hong, A.~Upfill-Brown,
  G.~Blumstein, E.~L. Lord, A.~N. Shamie, M.~{van der Schaar}, N.~F. SooHoo,
  D.~Y. Park, A risk calculator for the prediction of {C5} nerve root palsy
  after instrumented cervical fusion, World Neurosurgery 166 (2022) e703--e710.
\newblock \href {https://doi.org/https://doi.org/10.1016/j.wneu.2022.07.082}
  {\path{doi:https://doi.org/10.1016/j.wneu.2022.07.082}}.

\bibitem{sculley2015hidden}
D.~Sculley, G.~Holt, D.~Golovin, E.~Davydov, T.~Phillips, D.~Ebner,
  V.~Chaudhary, M.~Young, J.-F. Crespo, D.~Dennison, Hidden technical debt in
  machine learning systems, Advances in neural information processing systems
  28 (2015).

\bibitem{nestor2018rethinking}
B.~Nestor, M.~McDermott, G.~Chauhan, T.~Naumann, M.~C. Hughes, A.~Goldenberg,
  M.~Ghassemi, Rethinking clinical prediction: {W}hy machine learning must
  consider year of care and feature aggregation, Machine Learning for Health
  (ML4H) Workshop at NeurIPS (2018).

\bibitem{Cox1972}
D.~R. Cox, Regression models and life-tables, Journal of the Royal Statistical
  Society: Series B (Methodological) 34~(2) (1972) 187--202.
\newblock \href
  {https://doi.org/https://doi.org/10.1111/j.2517-6161.1972.tb00899.x}
  {\path{doi:https://doi.org/10.1111/j.2517-6161.1972.tb00899.x}}.

\bibitem{AKBILGIC2019}
O.~Akbilgic, R.~L. Davis, The promise of machine learning: {W}hen will it be
  delivered?, Journal of Cardiac Failure 25~(6) (2019) 484--485.
\newblock \href
  {https://doi.org/https://doi.org/10.1016/j.cardfail.2019.04.006}
  {\path{doi:https://doi.org/10.1016/j.cardfail.2019.04.006}}.

\bibitem{Schulz2020}
M.-A. Schulz, B.~T.~T. Yeo, J.~T. Vogelstein, J.~Mourao-Miranada, J.~N. Kather,
  K.~Kording, B.~Richards, D.~Bzdok, Different scaling of linear models and
  deep learning in {UKBiobank} brain images versus machine-learning datasets,
  Nature Communications 11~(1) (2020) 4238.
\newblock \href {https://doi.org/10.1038/s41467-020-18037-z}
  {\path{doi:10.1038/s41467-020-18037-z}}.

\bibitem{guyon2003introduction}
I.~Guyon, A.~Elisseeff, An introduction to variable and feature selection,
  Journal of machine learning research 3~(Mar) (2003) 1157--1182.

\bibitem{caruana2015intelligible}
R.~Caruana, Y.~Lou, J.~Gehrke, P.~Koch, M.~Sturm, N.~Elhadad, Intelligible
  models for healthcare: {P}redicting pneumonia risk and hospital 30-day
  readmission, In: Proceedings of the 21th ACM SIGKDD International Conference
  on Knowledge Discovery and Data Mining (2015) 1721--1730.

\bibitem{Winkler2019}
J.~K. Winkler, C.~Fink, F.~Toberer, A.~Enk, T.~Deinlein, R.~Hofmann-Wellenhof,
  L.~Thomas, A.~Lallas, A.~Blum, W.~Stolz, H.~A. Haenssle, {Association Between
  Surgical Skin Markings in Dermoscopic Images and Diagnostic Performance of a
  Deep Learning Convolutional Neural Network for Melanoma Recognition}, JAMA
  Dermatology 155~(10) (2019) 1135--1141.
\newblock \href {https://doi.org/10.1001/jamadermatol.2019.1735}
  {\path{doi:10.1001/jamadermatol.2019.1735}}.

\bibitem{Yoon2022}
C.~H. Yoon, R.~Torrance, N.~Scheinerman, Machine learning in medicine: {S}hould
  the pursuit of enhanced interpretability be abandoned?, Journal of Medical
  Ethics 48~(9) (2022) 581--585.
\newblock \href {https://doi.org/10.1136/medethics-2020-107102}
  {\path{doi:10.1136/medethics-2020-107102}}.

\bibitem{FDA2019}
{Food and Drug Administration and others}, Proposed regulatory framework for
  modifications to artificial intelligence/machine learning ({AI/ML})-based
  software as a medical device ({SaMD}) (2019).

\bibitem{Mourby2021}
M.~Mourby, K.~{\'O}~Cathaoir, C.~B. Collin, Transparency of machine-learning in
  healthcare: The {GDPR} {\&} {E}uropean health law, Computer Law {\&} Security
  Review 43 (2021) 105611.

\bibitem{Kattan2016}
M.~W. Kattan, K.~R. Hess, M.~B. Amin, Y.~Lu, K.~G. Moons, J.~E. Gershenwald,
  P.~A. Gimotty, J.~H. Guinney, S.~Halabi, A.~J. Lazar, A.~L. Mahar, T.~Patel,
  D.~J. Sargent, M.~R. Weiser, C.~Compton, members of~the AJCC Precision
  Medicine~Core, {American Joint Committee on Cancer} acceptance criteria for
  inclusion of risk models for individualized prognosis in the practice of
  precision medicine, CA: A Cancer Journal for Clinicians 66~(5) (2016)
  370--374.
\newblock \href {https://doi.org/https://doi.org/10.3322/caac.21339}
  {\path{doi:https://doi.org/10.3322/caac.21339}}.

\bibitem{Beam2020}
A.~L. Beam, A.~K. Manrai, M.~Ghassemi, Challenges to the reproducibility of
  machine learning models in health care, JAMA 323~(4) (2020) 305--306.
\newblock \href {https://doi.org/10.1001/jama.2019.20866}
  {\path{doi:10.1001/jama.2019.20866}}.

\bibitem{LeVeque2012}
R.~J. LeVeque, I.~M. Mitchell, V.~Stodden, Reproducible research for scientific
  computing: {T}ools and strategies for changing the culture, Computing in
  Science \& Engineering 14~(4) (2012) 13--17.
\newblock \href {https://doi.org/10.1109/MCSE.2012.38}
  {\path{doi:10.1109/MCSE.2012.38}}.

\bibitem{Milkowski2018}
M.~Mi{\l}kowski, W.~M. Hensel, M.~Hohol, Replicability or reproducibility? {O}n
  the replication crisis in computational neuroscience and sharing only
  relevant detail, Journal of Computational Neuroscience 45~(3) (2018)
  163--172.
\newblock \href {https://doi.org/10.1007/s10827-018-0702-z}
  {\path{doi:10.1007/s10827-018-0702-z}}.

\bibitem{Feurer2015}
M.~Feurer, A.~Klein, K.~Eggensperger, J.~Springenberg, M.~Blum, F.~Hutter,
  Efficient and robust automated machine learning, Advances in Neural
  Information Processing Systems 28 (2015) 2755–2763.

\bibitem{mice2011}
S.~van Buuren, K.~Groothuis-Oudshoorn, {mice}: {M}ultivariate imputation by
  chained equations in {R}, Journal of Statistical Software 45~(3) (2011)
  1–67.
\newblock \href {https://doi.org/10.18637/jss.v045.i03}
  {\path{doi:10.18637/jss.v045.i03}}.

\bibitem{stekhoven2011missforest}
D.~J. Stekhoven, P.~B{\"u}hlmann, {MissForest—non-parametric missing value
  imputation for mixed-type data}, Bioinformatics 28~(1) (2011) 112--118.
\newblock \href {https://doi.org/10.1093/bioinformatics/btr597}
  {\path{doi:10.1093/bioinformatics/btr597}}.

\bibitem{Jarrett2022}
D.~Jarrett, B.~C. Cebere, T.~Liu, A.~Curth, M.~van~der Schaar, {H}yper{I}mpute:
  {G}eneralized iterative imputation with automatic model selection, In:
  Proceedings of the 39th International Conference on Machine Learning 162
  (2022) 9916--9937.

\bibitem{Liu2013}
J.~Liu, A.~Gelman, J.~Hill, Y.-S. Su, J.~Kropko, {On the stationary
  distribution of iterative imputations}, Biometrika 101~(1) (2013) 155--173.
\newblock \href {https://doi.org/10.1093/biomet/ast044}
  {\path{doi:10.1093/biomet/ast044}}.

\bibitem{van2018flexible}
S.~Van~Buuren, Flexible imputation of missing data, CRC press, 2018.

\bibitem{wang2017batched}
Z.~Wang, C.~Li, S.~Jegelka, P.~Kohli, Batched high-dimensional bayesian
  optimization via structural kernel learning, In: Proceedings of the 34th
  International Conference on Machine Learning (2017) 3656--3664.

\bibitem{Lisha2018hyperband}
L.~Li, K.~Jamieson, G.~DeSalvo, A.~Rostamizadeh, A.~Talwalkar, Hyperband: {A}
  novel bandit-based approach to hyperparameter optimization, Journal of
  Machine Learning Research 18~(185) (2018) 1--52.

\bibitem{Crone2006}
S.~F. Crone, S.~Lessmann, R.~Stahlbock, The impact of preprocessing on data
  mining: {A}n evaluation of classifier sensitivity in direct marketing,
  European Journal of Operational Research 173~(3) (2006) 781--800.
\newblock \href {https://doi.org/https://doi.org/10.1016/j.ejor.2005.07.023}
  {\path{doi:https://doi.org/10.1016/j.ejor.2005.07.023}}.

\bibitem{lundberg2017unified}
S.~M. Lundberg, S.-I. Lee, A unified approach to interpreting model
  predictions, Advances in Neural Information Processing Systems 30 (2017).

\bibitem{Crabbe2021Simplex}
J.~Crabbe, Z.~Qian, F.~Imrie, M.~van~der Schaar, Explaining latent
  representations with a corpus of examples, Advances in Neural Information
  Processing Systems 34 (2021) 12154--12166.

\bibitem{Crabbe2020}
J.~Crabbe, Y.~Zhang, W.~Zame, M.~van~der Schaar, Learning outside the
  black-box: {T}he pursuit of interpretable models, Advances in Neural
  Information Processing Systems 33 (2020) 17838--17849.

\bibitem{streamlit}
Streamlit, https://streamlit.io/.

\bibitem{Sudlow2015}
C.~Sudlow, J.~Gallacher, N.~Allen, V.~Beral, P.~Burton, J.~Danesh, P.~Downey,
  P.~Elliott, J.~Green, M.~Landray, B.~Liu, P.~Matthews, G.~Ong, J.~Pell,
  A.~Silman, A.~Young, T.~Sprosen, T.~Peakman, R.~Collins, {UK Biobank}: {A}n
  open access resource for identifying the causes of a wide range of complex
  diseases of middle and old age, PLOS Medicine 12~(3) (2015) 1--10.
\newblock \href {https://doi.org/10.1371/journal.pmed.1001779}
  {\path{doi:10.1371/journal.pmed.1001779}}.

\bibitem{Adamska2015}
L.~Adamska, N.~Allen, R.~Flaig, C.~Sudlow, M.~Lay, M.~Landray, Challenges of
  linking to routine healthcare records in {UK Biobank}, Trials 16~(2) (2015)
  O68.
\newblock \href {https://doi.org/10.1186/1745-6215-16-S2-O68}
  {\path{doi:10.1186/1745-6215-16-S2-O68}}.

\bibitem{world2016global}
W.~H. Organization, et~al., Global report on diabetes, World Health
  Organization, 2016.

\bibitem{Bang2009}
H.~Bang, A.~M. Edwards, , A.~S. Bomback, C.~M. Ballantyne, D.~Brillon, M.~A.
  Callahan, S.~M. Teutsch, A.~I. Mushlin, L.~M. Kern, Development and
  validation of a patient self-assessment score for diabetes risk, Annals of
  Internal Medicine 151~(11) (2009) 775--783.
\newblock \href {https://doi.org/10.7326/0003-4819-151-11-200912010-00005}
  {\path{doi:10.7326/0003-4819-151-11-200912010-00005}}.

\bibitem{Lindstrom2003}
J.~Lindstr{\"o}{\"m}, J.~Tuomilehto, {The Diabetes Risk Score: A practical tool
  to predict type 2 diabetes risk}, Diabetes Care 26~(3) (2003) 725--731.

\bibitem{Hippisley-Cox2017Diabetes}
J.~Hippisley-Cox, C.~Coupland, Development and validation of {QDiabetes-2018}
  risk prediction algorithm to estimate future risk of type 2 diabetes: cohort
  study, BMJ 359 (2017).
\newblock \href {https://doi.org/10.1136/bmj.j5019}
  {\path{doi:10.1136/bmj.j5019}}.

\bibitem{cohen2013statistical}
J.~Cohen, Statistical power analysis for the behavioral sciences, Routledge,
  2013.

\bibitem{Nano2017ggt}
J.~Nano, T.~Muka, S.~Ligthart, A.~Hofman, S.~Darwish~Murad, H.~L. Janssen,
  O.~H. Franco, A.~Dehghan, Gamma-glutamyltransferase levels, prediabetes and
  type 2 diabetes: {A} {M}endelian randomization study, International Journal
  of Epidemiology 46~(5) (2017) 1400--1409.
\newblock \href {https://doi.org/10.1093/ije/dyx006}
  {\path{doi:10.1093/ije/dyx006}}.

\bibitem{IDF2013}
{International Diabetes Federation}, IDF Diabetes Atlas, 6th edn., 2013.

\bibitem{Gale2001}
E.~A. Gale, K.~M. Gillespie, Diabetes and gender, Diabetologia 44~(1) (2001)
  3--15.
\newblock \href {https://doi.org/10.1007/s001250051573}
  {\path{doi:10.1007/s001250051573}}.

\bibitem{ADA_website}
{American Diabetes Association}, https://diabetes.org/diabetes/a1c/diagnosis,
  {L}ast accessed: 18th August 2022.

\end{thebibliography}

\end{document}